\documentclass[10pt]{article}
\usepackage[preprint]{tmlr}

\usepackage{graphicx}
\usepackage{color}
\usepackage{hyperref}

\usepackage{caption}
\usepackage{subcaption}
\usepackage{amsmath}
\usepackage[capitalise]{cleveref}
\usepackage{xcolor}
\usepackage{booktabs}
\usepackage{multirow}

\usepackage{xcolor}
\hypersetup{
    colorlinks,
    linkcolor={blue!50!black},
    citecolor={blue!50!black},
    urlcolor={blue!50!black}
}

\begin{document}

\title{Counterfactual contrastive learning:\\robust representations via causal image synthesis}

\author{\name Mélanie Roschewitz \email m.roschewitz21@imperial.ac.uk \\ 
      \addr Imperial College London, UK 
      \AND
\name Fabio De Sousa Ribeiro \email f.de-sousa-ribeiro@imperial.ac.uk \\
      \addr Imperial College London, UK
      \AND
      \name Tian Xia \email t.xia@imperial.ac.uk \\\addr Imperial College London, UK
      \AND
      \name Galvin Khara \email galvin@kheironmed.com \\
      \addr Kheiron Medical Technologies, UK
      \AND
    \name Ben Glocker \email b.glocker@imperial.ac.uk \\
      \addr Imperial College London, UK \& Kheiron Medical Technologies, UK\\}

\maketitle 
\begin{abstract}
Contrastive pretraining is well-known to improve downstream task performance and model generalisation, especially in limited label settings. However, it is sensitive to the choice of augmentation pipeline. Positive pairs should preserve semantic information while destroying domain-specific information. Standard augmentation pipelines emulate domain-specific changes with pre-defined photometric transformations, but what if we could simulate realistic domain changes instead? In this work, we show how to utilise recent progress in counterfactual image generation to this effect. We propose CF-SimCLR, a counterfactual contrastive learning approach which leverages approximate counterfactual inference for positive pair creation. Comprehensive evaluation across five datasets, on chest radiography and mammography, demonstrates that CF-SimCLR substantially improves robustness to acquisition shift with higher downstream performance on both in- and out-of-distribution data, particularly for domains which are under-represented during training. 

All code is available at \url{https://github.com/biomedia-mira/counterfactual-contrastive}.

\end{abstract}

\section{Introduction}

The use of self-supervised learning for training medical imaging models holds great promise. Contrastive learning (CL) in particular has proven successful at tackling two major issues in the medical imaging domain: (i) sensitivity to domain shift and (ii) scarcity of high-quality annotated data, across various tasks and modalities \citep{azizi2021big,azizi2023robust,ghesu2022self,zhou2023foundation}. Despite its successes, one of the main challenges when applying CL to the medical imaging domain concerns the data augmentation pipeline, a critical design choice that has a major impact on the downstream model performance \citep{tian2020makes}. Notably, standard CL augmentation pipelines were designed for natural images and may be far from optimal for medical imaging. Herein, we hypothesise that with a better augmentation pipeline, targeted towards the modality at hand, we may substantially improve performance, particularly in terms of robustness to domain shifts. So, how can we generate better augmentations?
Recently, we have seen significant advances towards high-resolution, high-fidelity counterfactual image synthesis using deep generative models \citep{ribeiro2023high,fontanella2023diffusion}. Intuitively, a counterfactual image can be understood as a `what-if' image. For example, given a patient's chest radiograph, taken with scanner $A$, one may ask \textit{``what would the image have looked like if it had been acquired with scanner $B$?''}. In medical imaging, counterfactual inference models have so far mostly proven useful for test-time model interpretability \citep{augustin2022diffusion,sanchez2022healthy,atad2022chexplaining,matsui2022counterfactual,sun2023inherently}. While notable success in image generation and faithfulness to counterfactual conditioning has been reported~\citep{monteiro2023measuring,ribeiro2023high}, leveraging counterfactual images to directly improve downstream performance in clinically relevant tasks remains to be seen. 

\begin{figure}[t]
    \centering
    \includegraphics[width=.98\textwidth]{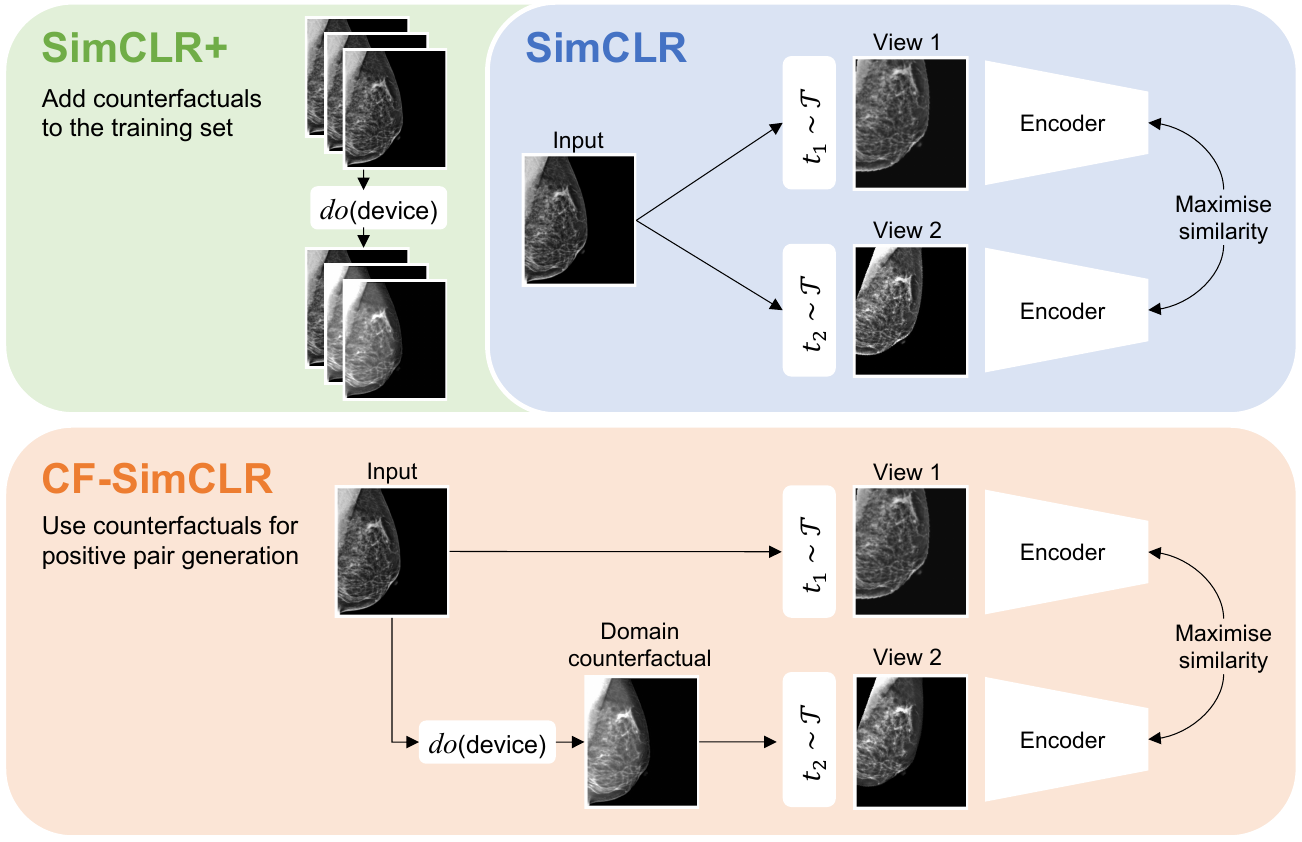}
    \caption{Summary of the three contrastive learning strategies compared in this study. The ``do'' operation denotes the counterfactual image generation model. $\mathcal{T}$ denotes the distribution of random augmentations applied to the images. In CF-SimCLR we systematically pair real and domain counterfactual images to generate cross-domain positive pairs. In SimCLR\texttt{+} , we simply add the counterfactuals to the training set but do not explicitly match real and counterfactual data during pair creation, i.e. they are independent training samples.}
    \label{fig:fig1}
\end{figure}

In this work, we investigate the use of counterfactual image generation in contrastive learning of image representations. Aiming to improve the robustness of learned representations, we introduce \emph{counterfactual contrastive learning}, where we combine real images and corresponding domain counterfactuals to construct positive pairs, thereby explicitly incorporating the counterfactual nature of the generated images into the contrastive learning objective. To isolate the effect of counterfactual positive pair generation, we compare our proposed counterfactual contrastive learning strategy to the case where counterfactual images are simply added to the training set, but not explicitly used in the contrastive loss (see \cref{fig:fig1}). Our strategy is integrated into the popular contrastive learning method SimCLR \citep{chen2020simple} and is compatible with many other contrastive learning methods.

Our results demonstrate that counterfactual contrastive learning substantially improves robustness to acquisition shift. Crucially, we find that this improvement stems from the explicit use of domain counterfactuals for positive pair generation. Our new pretraining strategy, called CF-SimCLR, is evaluated on two high-resolution image modalities, chest radiography and mammography, across five different datasets and challenging classification tasks.

\section{Related work}

\subsection{Contrastive learning} 
Contrastive learning consists of creating pairs of image \textit{views} that share semantic information (positive pairs), and using a contrastive objective to learn similar representations for these pairs whilst pushing them away from the representations of unrelated inputs (negative pairs). A seminal work in the CL field is SimCLR proposed by Chen \citet{chen2020simple}, which was later extended in various follow-up works e.g. BYOL \citep{grill2020bootstrap}, MoCo \citep{he2020momentum}, SWaV \citep{caron2020unsupervised}. SimCLR remains a simple, effective and widely used method, especially in medical imaging. \citet{azizi2023robust} showed that SimCLR pretraining improves model generalisation across multiple sources of dataset shifts. A critical choice in CL concerns the augmentations used for (artificially) generating the contrastive views. The design of the augmentation pipeline can substantially affect the characteristics of the learned representations, directly impacting the downstream performance and generalisation capabilities \citep{tian2020makes,scalbert2023improving}. So far, most standard augmentation pipelines originally designed for natural images have been used for medical imaging applications, which may be far from optimal for addressing the intricacies of the medical imaging domain. In medical imaging tasks, we often observe large intra-class differences due to differences in data acquisition which can overshadow inter-class differences such as subtle signs of disease. If not addressed, image representations obtained with self-supervised learning such as SimCLR may encode non-relevant differences due to acquisition shifts, which hampers downstream performance and robustness.

\subsection{Counterfactual image generation}
Counterfactual image generation methods can be broadly divided into two groups. First, methods focusing on interpretability, generating ``counterfactual explanations'' i.e. images that would reverse the decision of a given classifier while staying close to the original individual at test-time~\citep{augustin2022diffusion,sanchez2022healthy,atad2022chexplaining,matsui2022counterfactual,sun2023inherently}. The second set of methods focuses on generative models able to directly generate `what-if' images, independently of any downstream classifier, following principles of Pearl's ladder of causation \citep{pearl2009causality}. \citet{pawlowski2020deep} first introduced Deep Structural Causal Models (DSCM) to generate principled counterfactuals of small images. Recently, \citet{ribeiro2023high} extended the DSCM framework to high-resolution medical images, leveraging hierarchical variational autoencoders (HVAE) to significantly improve image quality. While there has been a surge of interest in these models, few studies have explored their usefulness for downstream applications e.g. data augmentation in imbalanced datasets \citep{xia2022adversarial,garrucho2023high}, or for improving fairness \citep{dash2022evaluating}. Therefore, the potential of counterfactual image generation models for improving downstream task performance remains largely under-explored.

\subsection{Combining contrastive learning and counterfactuals} In the vision-language literature, \citet{zhang2020counterfactual} proposed to leverage counterfactual text-image pairs in a CL objective for vision-language grounding tasks, where the creation of negative and positive counterfactuals is directly defined by the downstream task. In graph learning, \citet{yang2023generating} propose to leverage graph counterfactuals to generate hard negative examples for supervised CL. To the best of our knowledge, the usefulness of purely image-based counterfactuals in CL settings is yet to be explored.

\section{Counterfactual contrastive learning}

We introduce \emph{counterfactual contrastive learning} where the goal is to teach an image encoder to ignore domain-specific image characteristics using self-supervised learning and counterfactual inference. In standard contrastive learning, this is the role of photometric augmentations, aiming to teach the network to ignore contrast, brightness and other variations. However, such simple augmentations are unable to capture realistic variations caused by acquisition shifts (e.g., differences across scanners). Acquisition shift is caused by many complex factors which are difficult to model with handcrafted augmentation pipelines. We instead propose to use a counterfactual image model to generate realistic domain changes. Our investigation can be summarised in three steps: (i) training of a counterfactual image generation model; (ii) pretraining an encoder in a contrastive manner (with and without counterfactuals); (iii) evaluating the pretrained encoder with linear probing and finetuning on challenging downstream tasks and datasets. \cref{fig:fig1} summarises all compared contrastive pretraining strategies in this study.

\subsection{Counterfactual image generation}
To generate image counterfactuals, we build a DSCM following \citet{ribeiro2023high} by training an HVAE conditioned on the assumed causal parents of the image (see \cref{sec:cf_gen} for details on the causal graphs).
The model architecture mostly follows the one proposed by the authors for chest X-ray counterfactuals. To further improve the quality of the HVAE we added an embedding layer to encode raw parent variables prior to conditioning the HVAE and slightly modified the normalisation layers to improve training stability\footnote{All the code is made available at \url{https://github.com/biomedia-mira/counterfactual-contrastive}}. To avoid relying on external classifiers to improve generation quality (as they may introduce their own biases), we omitted the optional counterfactual finetuning step proposed in~\citet{ribeiro2023high}, simplifying the training pipeline of our HVAE.

\subsection{CF-SimCLR}
\label{sec:cfsimclr}

We use SimCLR~\citep{chen2020simple} as our simple yet effective CL strategy where positive pairs are created by applying non-deterministic transformations to every image, creating two distinct ``views'' that share the same semantic information. Every view is then passed through an image encoder (e.g. ResNet) to produce high dimensional embeddings, which are projected to a smaller dimension with an MLP. The contrastive loss is then applied to these projected representations $z$. For a batch of $N$ images, given one positive pair, all other $2(N-1)$ data points are treated as negative samples. For each positive pair, the SimCLR loss is defined as
$$\mathcal{L}_{i,j} = - \log \frac{\exp(\text{sim}(z_i, z_j)/\tau)}{\sum_{k=1, k\ne i}^{2N} \exp(\text{sim}(z_i, z_k)/\tau)},$$ where $\text{sim}(u,v) = \frac{u^{T}v}{||u||||v||}$. 
To compute the final loss we sum $\mathcal{L}_{i,j}$ over all positive pairs $i,j$ in the batch. \\

In CF-SimCLR, we propose to use domain counterfactuals to create cross-domain positive pairs, instead of only relying on random augmentations for pair generation. For a given image, we first generate a domain counterfactual (if several domains are available we sample one uniformly at random). Then, both images, the original one and its corresponding domain counterfactual, are fed through the random augmentation pipeline; the rest of the CL pipeline is identical to standard SimCLR, see \cref{fig:fig1}. SimCLR is widely used in the medical imaging domain~\citep{azizi2023robust}, however, we stress that our counterfactual contrastive learning framework is general and directly applicable to many other CL strategies.

\section{Experiments}
\subsection{Datasets} 
We compare three pretraining strategies across two modalities and five multi-source publicly available datasets. On chest radiography, we use PadChest~\citep{bustos2020padchest}, a large dataset with scans from two different scanners for pretraining. The availability of scanner information makes this dataset an ideal choice for pretraining and counterfactual generation. For evaluation, we focus on pneumonia detection. First, we evaluate on in-distribution (ID) test data from both PadChest scanners. To evaluate how well the features transfer to out-of-distribution (OOD) domains, we use the RSNA Pneumonia Detection~\citep{rsna} and CheXpert~\citep{irvin2019chexpert} datasets. Secondly, we test our approach on mammography, focusing on breast density prediction (imbalanced 4-class classification). We use EMBED~\citep{jeong2022emory}, a large dataset collected in the US using six different scanners. We use data from five scanners for pretraining (referred to as `EMBED' in the following) and keep the remaining scanner as a hold-out domain for OOD evaluation (i.e. `Senographe Essential'). Note that in EMBED, more than 90\% of the data comes from a single scanner, which allows us to assess model performance on under-represented domains only available in small proportions at (pre-)training time. For additional OOD transfer evaluation, we use VinDR-Mammo~\citep{Nguyen2022} (two scanners, collected in Vietnam). See \cref{tab:dataset_splits} in the Appendix for details on dataset sizes, splits and inclusion criteria.

\subsection{Counterfactual generation}
\label{sec:cf_gen}

\begin{figure}
  \centering
  \hfill
  \begin{subfigure}[t]{0.6\textwidth}
  \centering
    \includegraphics[height=9cm]{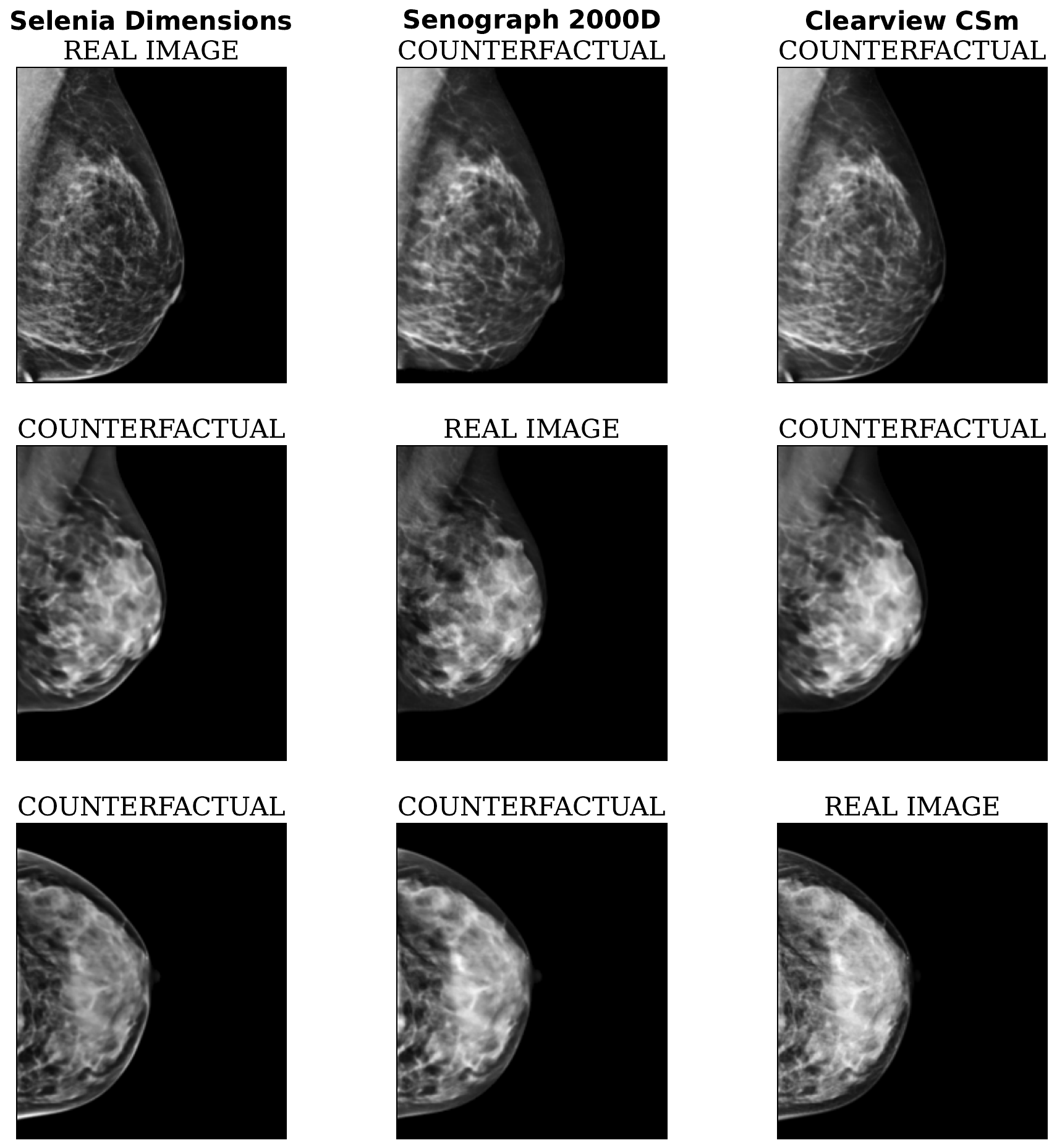}
    \caption{EMBED}
  \end{subfigure}
  \hfill
  \begin{subfigure}[t]{0.39\textwidth}
  \centering
    \includegraphics[height=9cm]{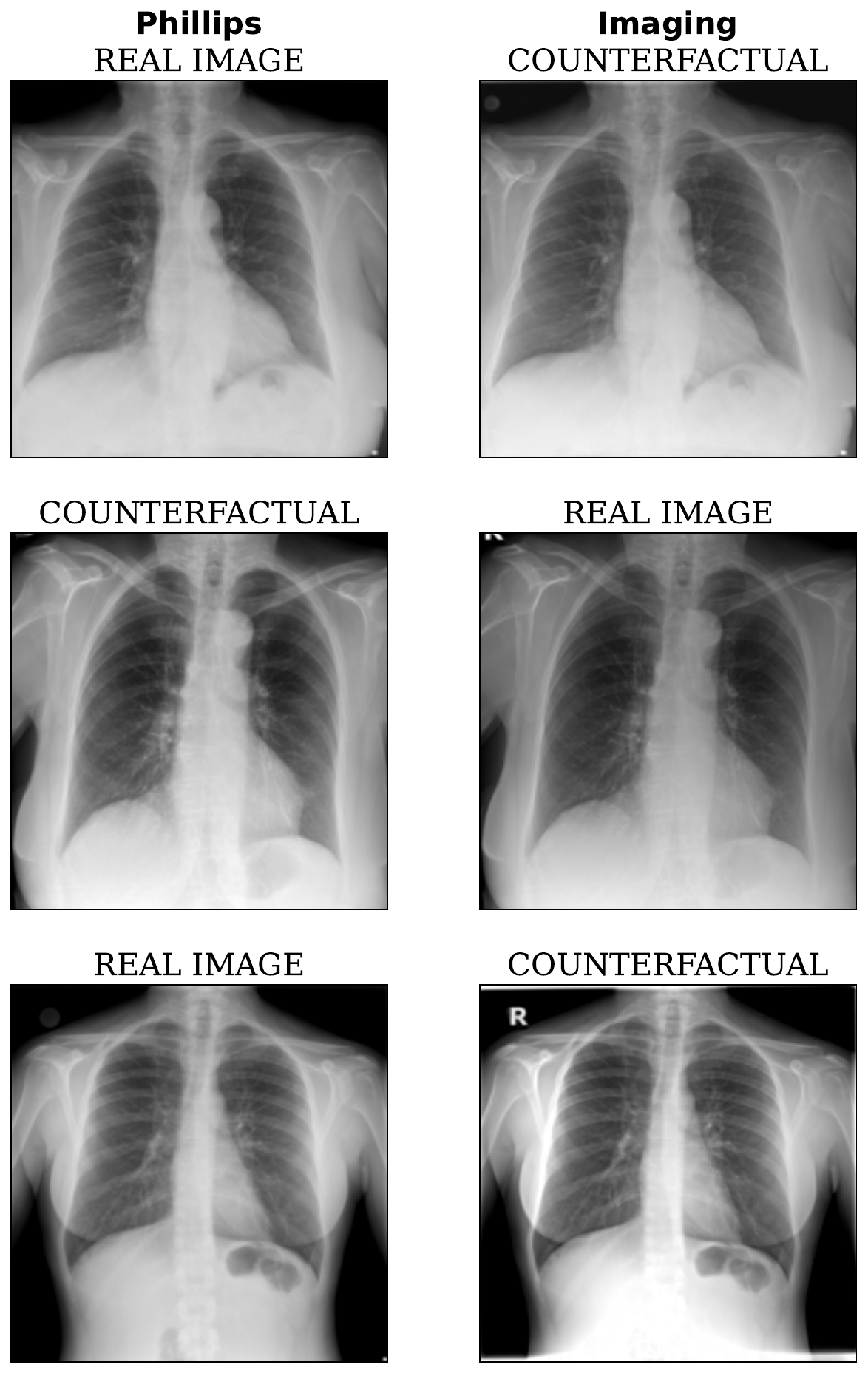}
    \caption{PadChest}
  \end{subfigure}
  \hfill
  \caption{Examples of counterfactual images generated using our model. Note that on PadChest, text is only imprinted on a subset of Imaging scans (not on Phillips), and our counterfactual model respects this by removing text when generating counterfactuals from Imaging to Phillips and vice-versa. Generated images have a resolution of 224x224 pixels for PadChest, 224x192 for EMBED.}
  \label{fig:cf_viz}
\end{figure}

To train our counterfactual inference model, the causal graph describing the data generating process must be defined. For many applications, such as interpretability, causal graphs would usually be chosen to faithfully reflect physiological and imaging processes. In this work, we focus on generating domain counterfactuals i.e. we only intervene on the `scanner' indicator variable. Hence, `scanner' is the only variable that is explicitly required in the causal graph, and all other factors of variation will be captured in the exogenous noise during counterfactual inference. Importantly, the assumed causal graphs do not include any parents relative to the downstream task (such as disease labels), as this would violate the \emph{unsupervised} assumptions of contrastive learning. For EMBED, we only included `scanner' as a parent in the DSCM, in PadChest we included `sex' and `scanner' (`sex' being optional). We show some qualitative examples of generated domain counterfactuals in \cref{fig:cf_viz}. In terms of \emph{effectiveness}~\citep{monteiro2023measuring}, the generated domain counterfactuals on PadChest are able to fool a domain classifier trained on real data 95\% of the time, whereas on EMBED the model can fool the domain classifier 80\% of the time when generating counterfactuals uniformly at random across all devices. Note that 4 out of 5 scanners represent less than 5\% of the EMBED training set, thereby increasing the difficulty of counterfactual generation on these underrepresented scanners. To help counter this imbalance during training we used weighted batch sampling.

\subsection{Contrastive pretraining and evaluation} All encoders use a ResNet-50 architecture, initialised with ImageNet weights and we use a 512 batch size for CL pretraining. To ensure fair comparisons, we kept all hyperparameters constant across various pretraining strategies. Following standard evaluation practices, we compare pretrained encoders with linear probing (frozen encoder) and full model finetuning with varying amounts of labelled data, using a weighted cross-entropy loss. All models are finetuned with real data only. For both tasks, we evaluate models on ID data (same dataset as for pretraining), as well as on external datasets. Evaluation of external datasets allows us to determine whether counterfactual image generation is useful even when the finetuning domain is disjoint from the pretraining domains and outside of the distribution of generated counterfactuals.

\section{Results}

\begin{figure}%
    \centering
    \includegraphics[width=\textwidth]{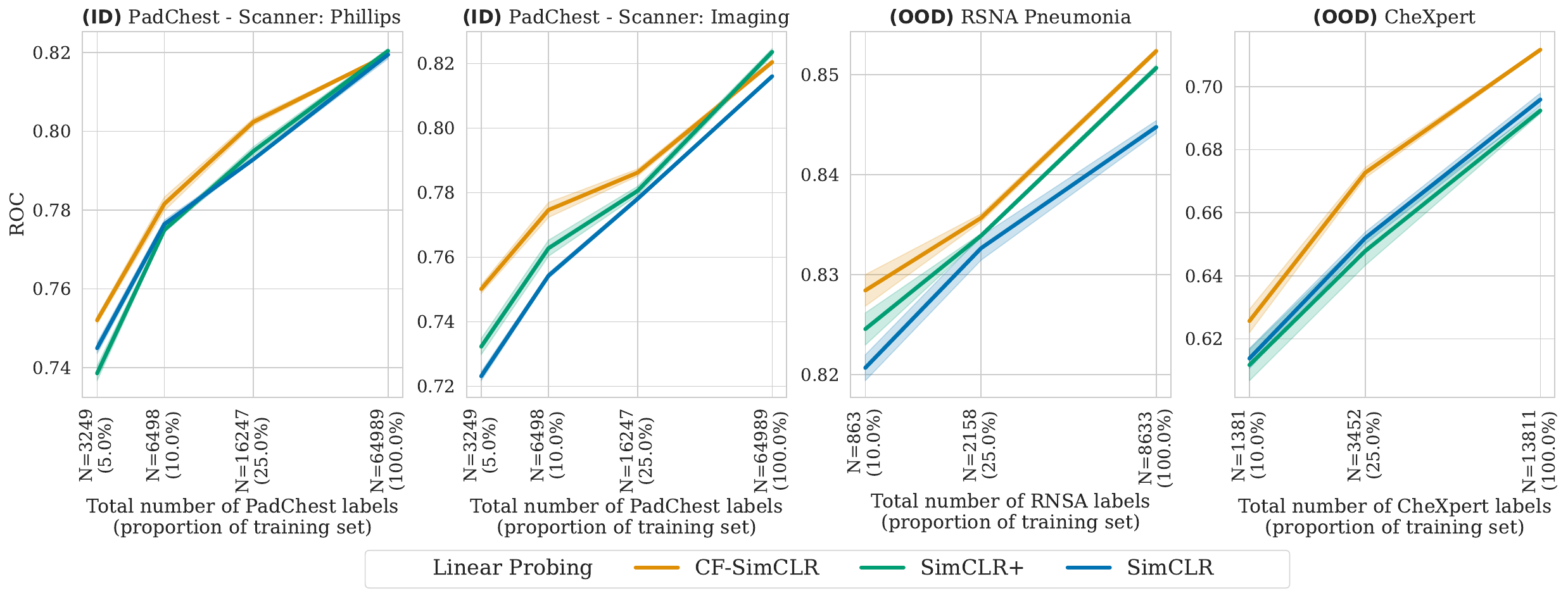}
    \caption{Pneumonia detection results with linear probing  (frozen encoder), reported as average ROC-AUC over 3 seeds, shaded areas denote +/- one standard error. CF-SimCLR consistently outperforms encoders trained with standard SimCLR and SimCLR\texttt{+} where counterfactuals are added to the training set.}
    \label{fig:xray_lin}
\end{figure}

\begin{figure}
    \centering
    \includegraphics[width=\textwidth]{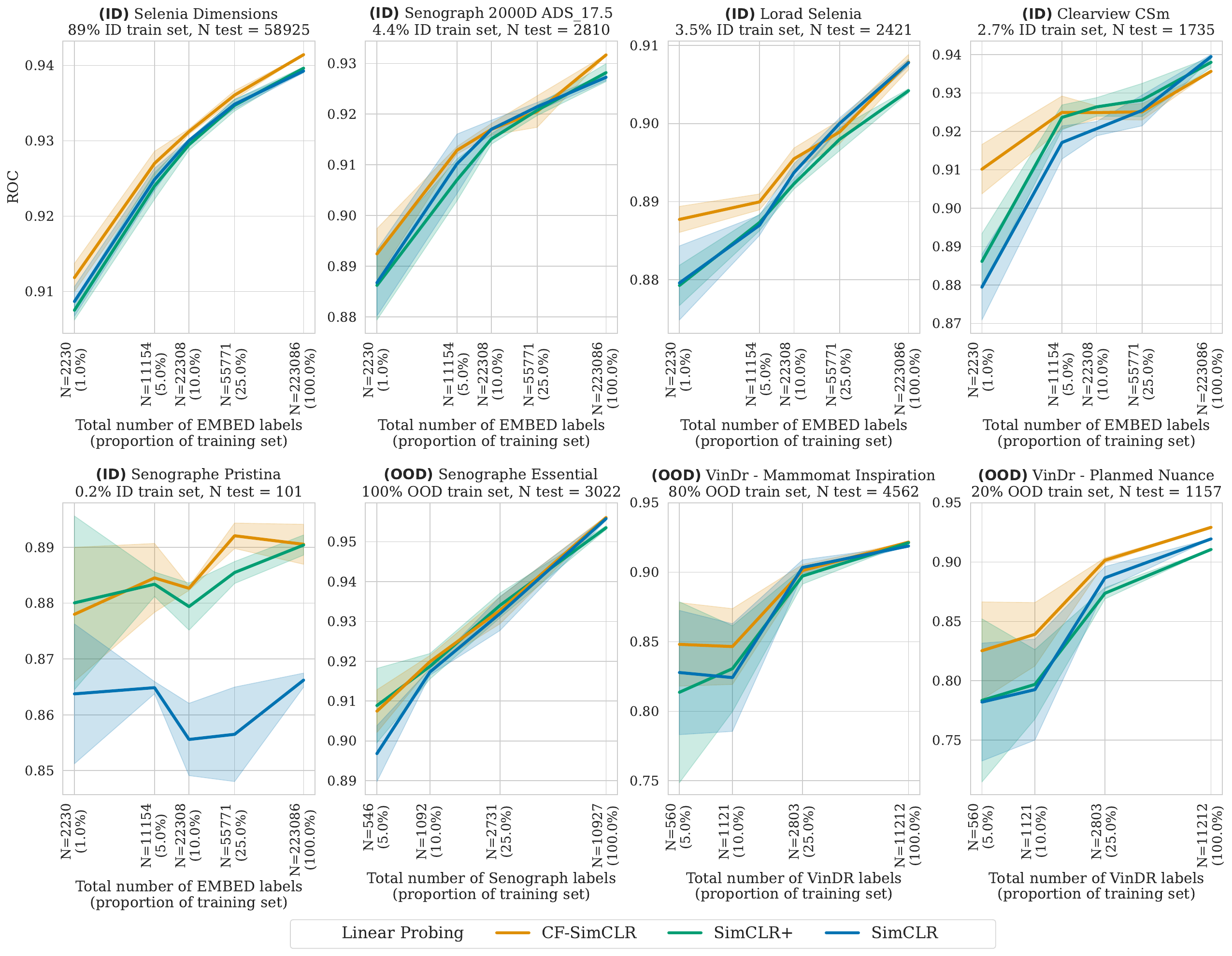}
    \caption{Breast density results with linear probing (frozen encoder), reported as average one-versus-rest macro ROC-AUC over 3 seeds, shaded areas denote +/- one standard error. CF-SimCLR overall performs best across ID and OOD data.}
    \label{fig:mammo_lin}
\end{figure}

 In \cref{fig:xray_lin} and \cref{fig:mammo_lin}, we present linear probing results for 3 pretraining strategies: (i) standard SimCLR, (ii) SimCLR\texttt{+} with a training set augmented with counterfactuals, (iii) CF-SimCLR. The difference between (ii) and (iii) lies in the construction of positive pairs. In SimCLR\texttt{+}, we simply add the counterfactuals to the training set but do not explicitly match real and counterfactual data during pair creation, i.e. they are independent training samples. In CF-SimCLR we systematically pair real and domain counterfactual images as per \cref{fig:fig1}.

\paragraph{Do counterfactuals improve the quality and robustness of contrastively learned representations?}
Results on pneumonia detection (\cref{fig:xray_lin}) show that, on all datasets, CF-SimCLR improves downstream performance compared to standard SimCLR (orange versus blue), across all levels of label availability. Importantly, performance gains are substantial on external datasets, even though these images were outside of the pretraining (and counterfactual generation) domains. On mammography, linear probing results in \cref{fig:mammo_lin} show that CF-SimCLR consistently outperforms standard SimCLR across all ID scanners, in all scenarios with up to 20k labelled images. Importantly, for limited labels (<20k), CF-SimCLR performance gains are highest for under-represented scanners (all except Selenia Dimensions). In cases with large amounts of labelled data (>50k), encoders perform similarly for most scanners, except for Selenia Dimensions and Senographe Pristina where CF-SimCLR again surpasses SimCLR. On the external VinDR dataset, for limited labels, CF-SimCLR beats both baselines on both scanners. Importantly, on the under-represented OOD scanner (PlanMed Nuance) CF-SimCLR performs better by a substantial margin. Findings are similar for model finetuning with varying amounts of data (see Appendix \cref{fig:xray,fig:mammo}).

 \begin{figure}
    \centering
    \includegraphics[width=\textwidth]{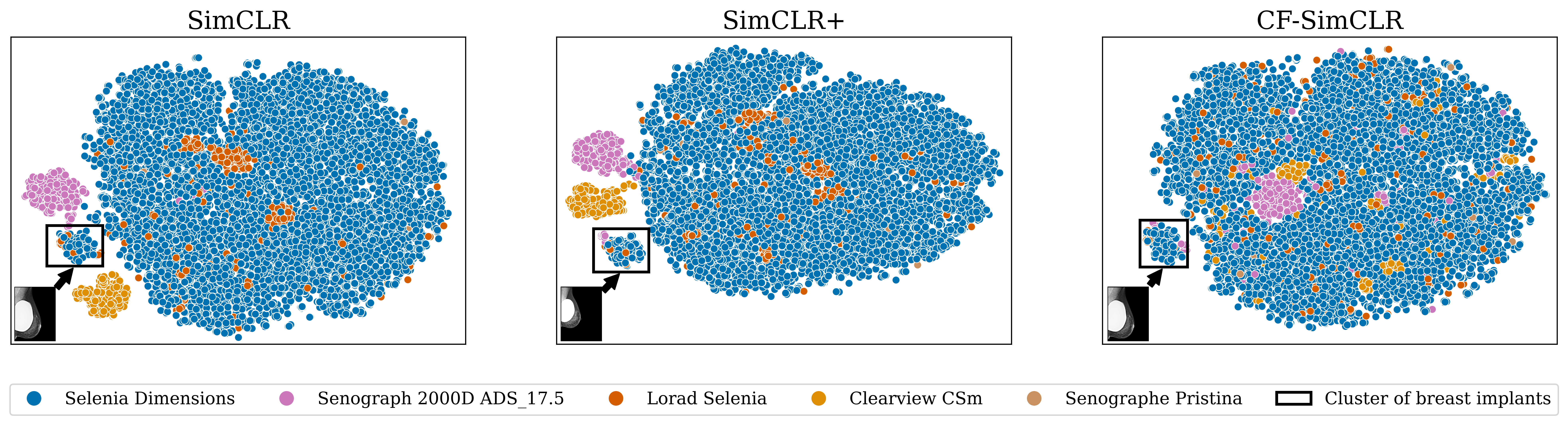}
    \caption{t-SNE projections of embeddings from mammography encoders. Encoders trained with standard SimCLR and SimCLR\texttt{+} exhibit domain clustering. Our CF-SimCLR embeddings are substantially less domain-separated and the only disjoint cluster exclusively contains breasts with implants, semantically different. Thumbnails show one randomly sampled image from each `implant' cluster.}
    \label{fig:mammo-embeddings}
\end{figure}

\paragraph{Is it advantageous to incorporate image counterfactuals in the contrastive objective or is it sufficient to simply add counterfactual data to the training set?} We find that CF-SimCLR performed consistently better than SimCLR\texttt{+} where the model is trained on the augmented training set, across datasets and most levels of labels. While SimCLR\texttt{+} sometimes improves over standard SimCLR, performance gains are not consistent and not as large as for models trained with CF-SimCLR (orange vs green in \cref{fig:xray_lin,fig:mammo_lin}). Differences were especially noticeable on under-represented scanners and for transfer to OOD datasets. This aligns with the theory: the counterfactual contrastive objective encourages explicit alignment of domains in the learned representation. An increase in domain alignment is in turn expected to improve robustness to acquisition shift, i.e. improve performance on under-represented scanners and for transfer learning with limited data. We also visualise embeddings of 16,000 randomly sampled test images for each mammography encoder in \cref{fig:mammo-embeddings}, showing that models trained with SimCLR as well as SimCLR\texttt{+} depict very clear domain separation in their t-SNE plots, whereas for CF-SimCLR embeddings appear much less domain-separated.

\paragraph{What about the computational overhead?} CF-SimCLR requires training a counterfactual inference model. We want to highlight that the generation model used in this work is lightweight compared to other image generation models (e.g. diffusion models). Taking EMBED as an example, we were able to generate >1M images in less than 7 hours on a single NVIDIA 3090 GPU, requiring 20GB of VRAM. Importantly, model training was fast (20 epochs were sufficient to yield good results). This is low compared to the computational cost of the contrastive learning part: to train SimCLR we required more powerful GPUs (2x NVIDIA L40 with 46GB VRAM each) and needed to train the model for 450 epochs.

\section{Conclusion}

We propose CF-SimCLR, a counterfactual contrastive learning method to improve the robustness of learned representations to acquisition shift. Extensive evaluation shows that incorporating counterfactual image generation in the contrastive learning objective substantially improves downstream performance of pretrained models across modalities and domains; especially for under-represented domains. Crucially, these performance gains hold even when transferring to domains not seen during pretraining. Future work could extend CF-SimCLR to other types of dataset shifts, e.g. enforcing subgroup invariance to improve model fairness.

\vspace{1cm}
\section*{Acknowledgements}
M.R. is funded through an Imperial College London President's PhD Scholarship. F.R., T.X. and B.G. received funding from the European Research Council (ERC) under the European Union’s Horizon 2020 research and innovation programme (grant agreement No 757173, project MIRA, ERC-2017-STG). B.G. also received support from the Royal Academy of Engineering as part of his Kheiron/RAEng Research Chair in Safe Deployment of Medical Imaging AI. We acknowledge support of the UKRI AI programme, and the Engineering and Physical Sciences Research Council, for CHAI - EPSRC AI Hub for Causality in Healthcare AI with Real Data (grant number EP/Y028856/1).

\newpage
\bibliography{bib}
\bibliographystyle{tmlr}

\newpage
\appendix
\setcounter{table}{0}
\renewcommand{\thetable}{\Alph{section}\arabic{table}}
\setcounter{figure}{0}
\renewcommand{\thefigure}{\Alph{section}\arabic{figure}}

\section{Datasets splits and inclusion criteria}

\begin{table}[h!]
    \centering
    \caption{Datasets splits and inclusion criteria. Splits are created at the patient level. $^{(*)}$ excluding Senographe Essential, kept as separate hold-out domain. Instructions to download all datasets and reproduce all experiments can be found in our public codebase \url{https://github.com/biomedia-mira/counterfactual-contrastive}.}
    \begin{tabular}{lcccc}
        \toprule
         \multirow{2}{*}{Dataset}  & \multirow{2}{*}{Inclusion criteria} & \multicolumn{3}{c}{Number of images} \\
         & & Train & Validation & Test \\
         \toprule
        EMBED~\citep{jeong2022emory} & 2D only$^{(*)}$ &   223,086 &  8,295 & 65,992 \\
        Senographe Essential~\citep{jeong2022emory} & - &   10,927 &  1,251 & 3,022 \\
        VinDR Mammo~\citep{vindr} & - & 11,191 & 2,813 & 5,996 \\
        PadChest~\citep{bustos2020padchest} & Adult PA only & 64,989 & 7,203 & 17,993 \\
        CheXpert~\citep{irvin2019chexpert} & PA only & 13,811 & 2,449 & 10,838 \\
        RSNA Pneumonia~\citep{rsna} & PA only & 8,633 & 1,524 & 4,354 \\
        \bottomrule
    \end{tabular}
    \label{tab:dataset_splits}
\end{table}

\section{Finetuning results}
\begin{figure}[htpb]
    \centering
    \includegraphics[width=.97\textwidth]{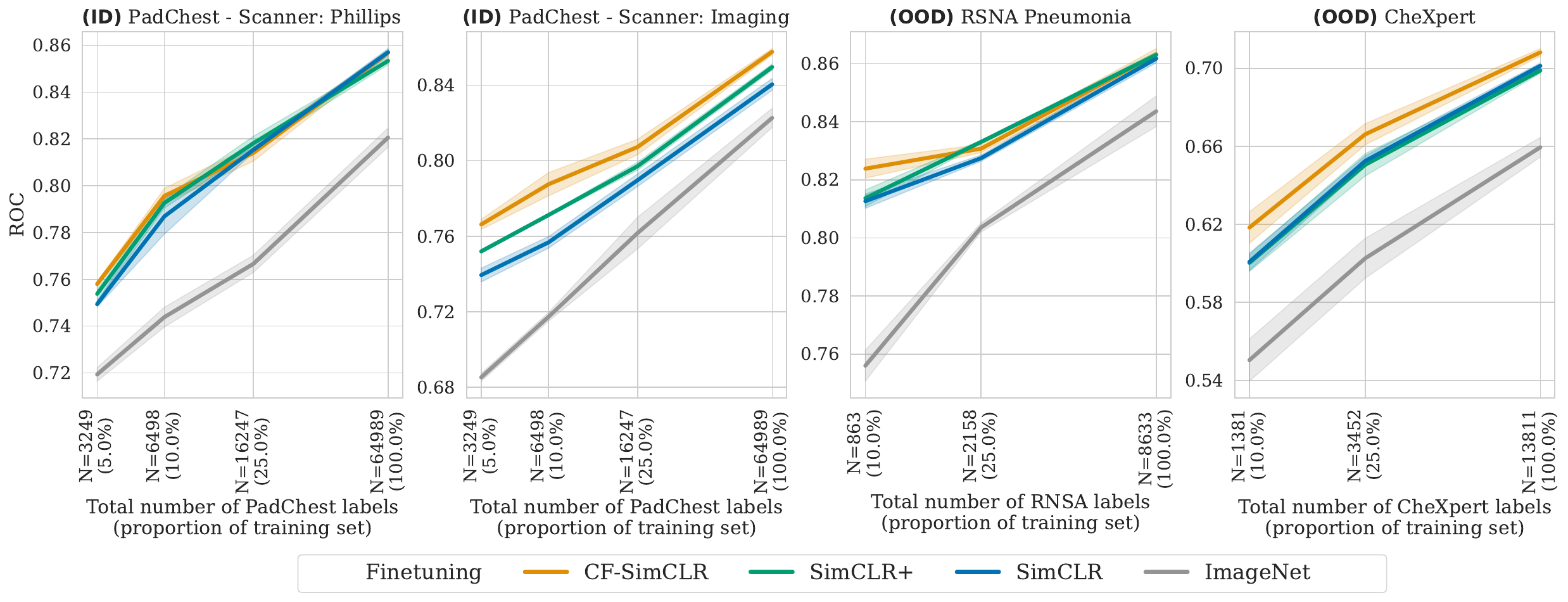}
    \caption{Pneumonia detection results with finetuning (unfrozen encoder). Reported as average ROC-AUC over 3 seeds, shaded areas denote +/- standard error. We also compare to a supervised baseline initialised with ImageNet weights. CF-SimCLR performs best overall ID and OOD.}
    \label{fig:xray}
\end{figure}

\begin{figure}[h!]
    \centering
    \includegraphics[width=.97\textwidth]{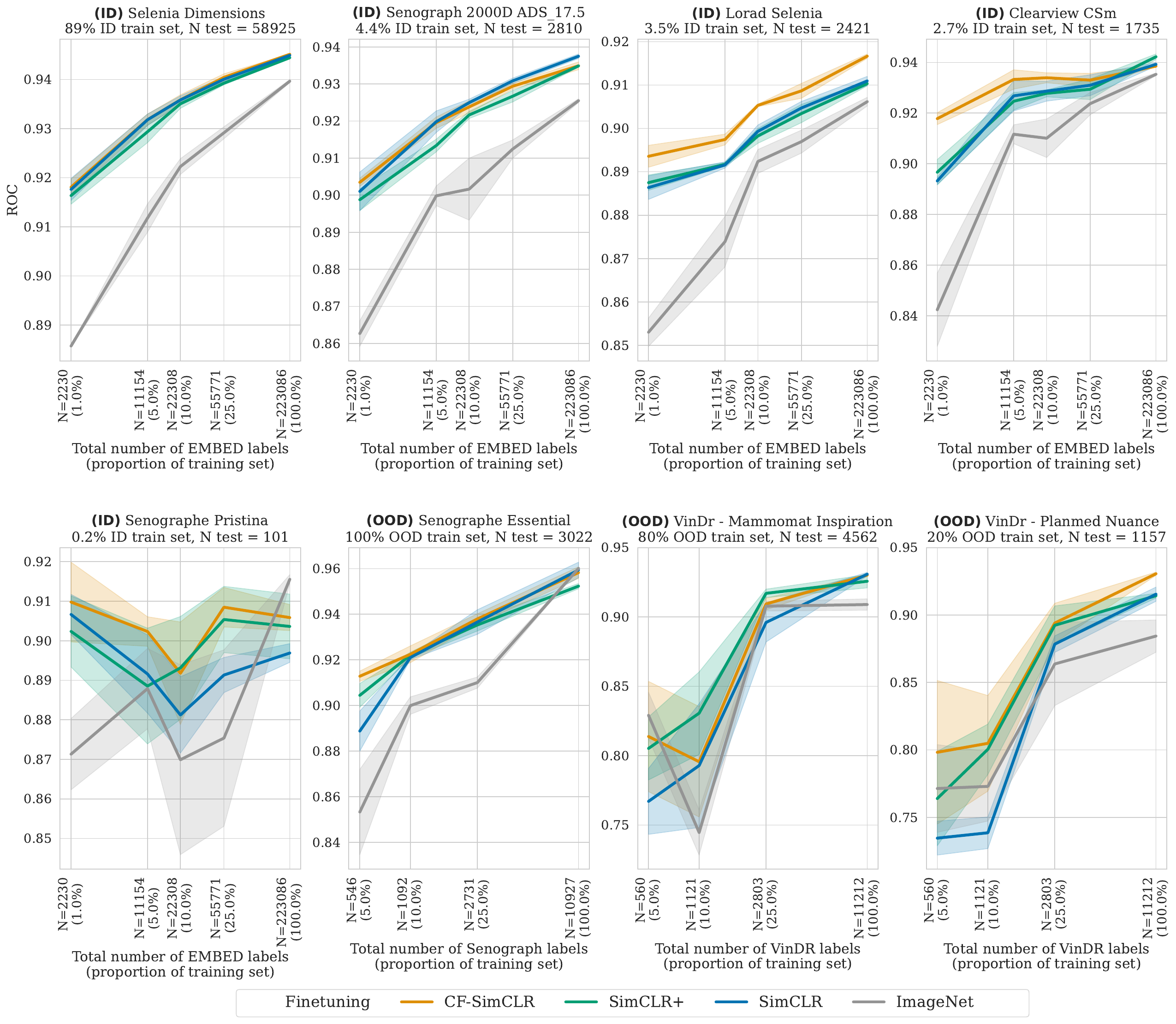}
    \caption{Breast density results with finetuning (unfrozen encoder), reported as average one-versus-rest macro ROC-AUC over 3 seeds, shaded areas denote +/- standard error. We also compare to a supervised baseline initialised with ImageNet weights. CF-SimCLR performs best overall ID and OOD.}
    \label{fig:mammo}
\end{figure}

\end{document}